\documentclass[journal]{IEEEtran}
\ifCLASSINFOpdf
\else
\fi
\usepackage{times}
\usepackage{epsfig}
\usepackage{graphicx}
\usepackage{amsmath}
\usepackage{amssymb}
\usepackage{booktabs}
\usepackage{etoolbox}
\usepackage{amssymb}
\usepackage{float}
\usepackage{cite}
\usepackage{multirow}
\usepackage{booktabs}
\usepackage{microtype}
\usepackage{setspace}
\usepackage{color}
\usepackage{threeparttable}
\usepackage{lineno}
\usepackage{amssymb}
\usepackage{bbm}
\usepackage{bbding}
\usepackage{pifont}
\usepackage{wasysym}
\usepackage{amssymb}
\usepackage{amssymb}
\begin{document}

\title{Edge-guided and Class-balanced Active Learning for Semantic Segmentation of Aerial Images}

\author{
        Lianlei Shan ~\IEEEmembership{Graduate Student member}, Weiqiang Wang\IEEEauthorrefmark{1} ~\IEEEmembership{Member},\\
        Ke Lv ~\IEEEmembership{Senior Member}, Bin Luo ~\IEEEmembership{Senior Member,~IEEE}
       
\thanks{* corresponding author

W. Wang and L. Shan are with the School
of Computer Science and Technology, University of Chinese Academy of Sciences, Beijing 100049, China
 (e-mail: shanlianlei18@mails.ucas.edu.cn, wqwang@ucas.ac.cn).
 
Ke Lv is with School of Engineering Science, University of Chinese Academy of Sciences, Beijing 100049, China 

Bin Luo is with MOE Key Lab of Signal Processing and Intelligent Computing, School of Computer Science and Technology, Anhui University, Hefei 230601, China
 }
 }

\maketitle
\begin{abstract}
Semantic segmentation requires pixel-level annotation, which is time-consuming. Active Learning (AL) is a promising method for reducing data annotation costs.
Due to the gap between aerial and natural images, the previous AL methods are not ideal, mainly caused by unreasonable labeling units and the neglect of class imbalance.
Previous labeling units are based on images or regions, which does not consider the characteristics of segmentation tasks and aerial images, i.e., the segmentation network often makes mistakes in the edge region, and the edge of aerial images is often interlaced and irregular.
Therefore, an edge-guided labeling unit is proposed and supplemented as the new unit.
On the other hand, the class imbalance is severe, manifested in two aspects: the aerial image is seriously imbalanced, and the AL strategy does not fully consider the class balance. Both seriously affect the performance of AL in aerial images.
We comprehensively ensure class balance from all steps that may occur imbalance, including initial labeled data, subsequent labeled data, and pseudo-labels. Through the two improvements, our method achieves more than 11.2\% gains compared to state-of-the-art methods on three benchmark datasets, Deepglobe, Potsdam, and Vaihingen, and more than 18.6\% gains compared to the baseline.
Sufficient ablation studies show that every module is indispensable.
Furthermore, we establish a fair and strong benchmark for future research on AL for aerial image segmentation.
\end{abstract}


\IEEEpeerreviewmaketitle

\section{Introduction}
\IEEEPARstart{W}ith the rapid development of satellites and aerial cameras, a large number of high-quality aerial images can be accessed, and how to cheaply obtain valuable information from these aerial images is an important research topic. Among them, semantic segmentation \cite{shan2021class,shan2021decouple,shan2021densenet,shan2021uhrsnet,shan2022class,shan2022mbnet,shan2023boosting,shan2023data,shan2023incremental,wu2023continual}, a pixel-level classification task, is one of the important and basic tasks. Semantic segmentation contains important application value in farmland area detection, disaster assessment, road extraction, and other aspects.
For the segmentation task of aerial images, there have been a lot of excellent works to optimize the segmentation models from the aspects of increasing receptive field \cite{tgrs2,tgrs3}, spatial relation \cite{tgrs_ccanet}, and valuable context \cite{tgrs_dense,tgrs1,zhao2023explore,zhao2023flowtext,zhao2023generative,zhao2024controlcap}.
However, the aforementioned models require pixel-level annotations during training.
And even with the help of some auxiliary annotation tools, pixel-level annotation still requires a lot of manpower and is time-consuming.
Active Learning (AL) is a suitable strategy to solve this problem because it can make networks achieve competitive performance with fewer labeled training data \cite{acl_survey}.

The process of AL is iterative \cite{activelearning}.
Firstly, some data is acquired and labeled as the initial labeled data.
Next, the labeled data is used to train the network. Then, the subsequent new data that need to be labeled is acquired according to the output of the trained network.
The data acquisition and network training are repeated until the network performance meets the requirements or the labeled data reaches a budget (10\% of the whole data).
However, due to the huge gap between aerial images and natural images, the AL method that achieves impressive results in natural images is not satisfactory in aerial images.
This is mainly due to the unreasonable labeling units and the neglect of class imbalance.
We analyze the two, respectively.

\begin{figure}[h]
\centering
\includegraphics[scale=0.265]{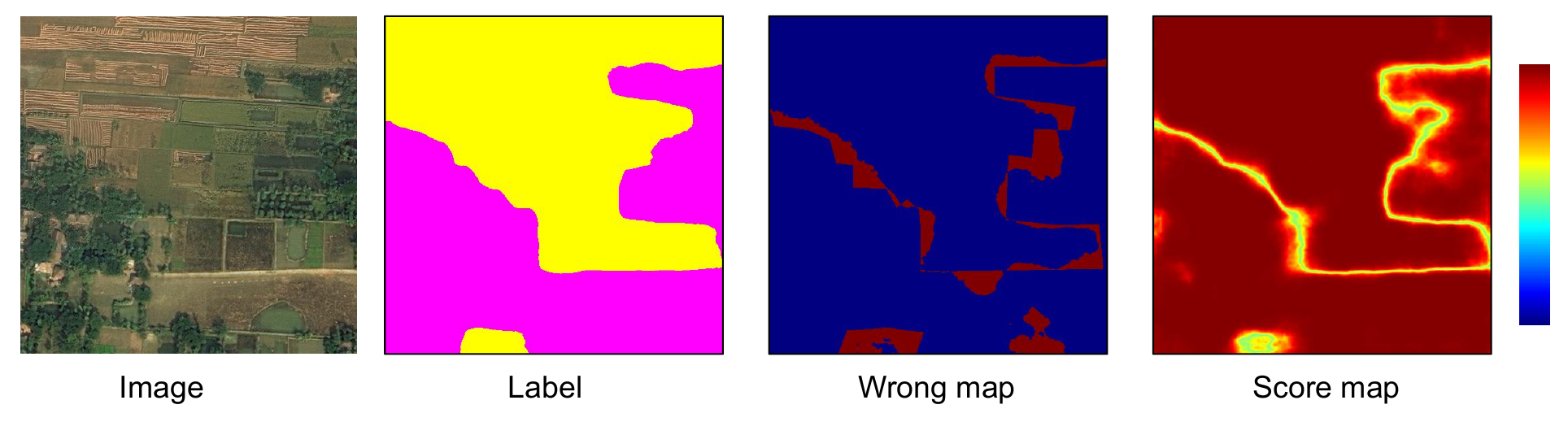}
\caption{
The image, label, wrong map, and score map are from left to right.
The error regions generally surround the edges, and the scores of the network output in the edge regions are also low (the uncertainty is large).
}
\label{pic_intro_score}
\end{figure}

\textbf{Labeling Unit:}
The labeling unit is the smallest labeling region, i.e., the unit is either not labeled or fully labeled. Previously, the labeling unit is image-based \cite{CBAL} or region-based \cite{spal,ralis}, i.e., labeling specific images or specific regions in images. Region-based methods are further divided into regular region-based methods \cite{ralis} and irregular region-based methods \cite{spal}. Regular regions are generally rectangles, while irregular regions are generally superpixels.
However, none of them consider the characteristics of the segmentation task and aerial images.
Fig. \ref{pic_intro_score} shows the regions with the wrong prediction, and it can be seen that the errors in the segmentation task are prone to appear in the edge regions.
Moreover, the score of the edge regions is also lower, and lower scores indicate uncertainty. Uncertainty contains information \cite{activelearning}, which is useful for network training \cite{spal}. Therefore, the edge regions deserve more attention. However, the previous rectangle-based and super-pixel-based methods do not consider the edge regions especially, which causes the edge regions to be scattered into rectangles or superpixels. Rectangle and superpixels already contain many correctly segmented results, resulting in a waste of annotation.

Based on the above analysis, we take the edge region alone as labeling units for the supplement. Specifically, we use traditional image processing algorithms to obtain edges and then diffuse to the surroundings to obtain edge regions. This approach can significantly reduce the number of manual annotations on the premise of ensuring network training.

\begin{figure}[h]
\centering
\includegraphics[scale=0.58]{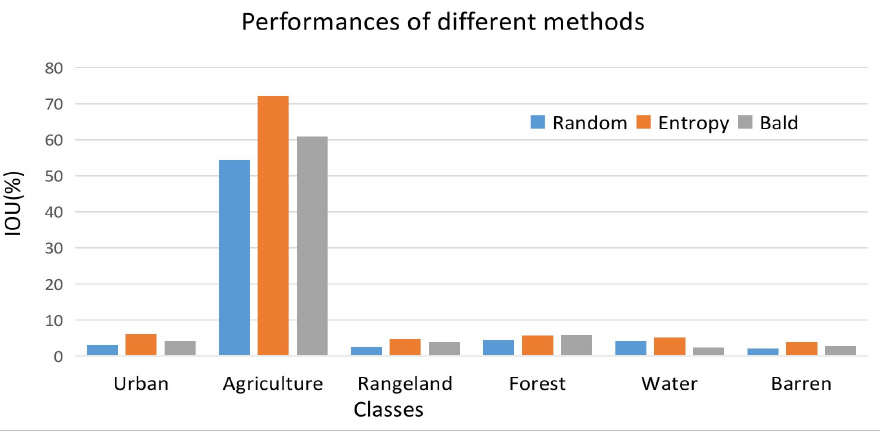}
\caption{
The results of different active learning methods on Deepglobe dataset \cite{deepglobe} with the labeled data accounting for 20\% of the total, and the segmentation network is Deeplabv3 \cite{deeplabv3}. 
In all three approaches, there are widespread performance imbalances between classes.
}
\label{pic_intro1}
\end{figure}

\textbf{Class Balance:}
The other factor affecting AL's performance in aerial images is class imbalance. Imbalance is reflected in: the original data is unbalanced; AL's acquisition strategy imbalanced the labeled data. Figure \ref{pic_intro1} shows the results for the commonly used AL methods. 
It can be seen that the model works well for class agriculture but is very poor for others, showing a severe class imbalance.
In fact, the class imbalance problem is also noticed by previous methods, and some methods are also proposed.
SpAL \cite{spal} uses superpixels instead of rectangular boxes as the basic unit of annotation and makes the range of a superpixel contain as many classes as possible to keep the data balance of each category.
RALIS \cite{ralis} employs reinforcement learning to optimize the IoU for each class, which is a commonly used balancing method.
CBAL \cite{CBAL} takes class balance directly as the optimization goal.
However, these methods do not completely consider the imbalance, which leads to the class imbalance still existing and seriously affects the role of AL in segmentation.

SpAL \cite{spal} keeps each category's sample size as balanced as possible. But some categories, such as water, have unique texture and color characteristics, so they do not need many samples to get satisfactory results. Therefore, we abandon the previous sample-based balance and use performance-based balance, i.e., giving more labeled samples to hard-to-learn categories.
Moreover, previous works ignore the selection of initial labeled data, leading to a class imbalance in the initial training. The initial imbalance is amplified in subsequent iterative acquisitions, creating a vicious circle.
Therefore, we design a specific strategy for acquiring the initial labeled data and maximize the use of the existing pre-trained networks to make the initial labeled data class-balanced.

We also quantify the high-confidence output of the network as the pseudo labels to participate in network training. And a class-specific and balanced threshold method is proposed to ensure the balance of pseudo-labels.
Besides, according to the characteristics that edge regions contain multiple categories, contrastive learning \cite{contrastiveseg} is introduced.
Unlike previous, a new feature selection strategy is proposed for contrastive learning so that the hard-to-learn categories can get more attention, thus ensuring class balance at the feature extraction level.
In a nutshell, we analyze all stages of data acquisition and add class balance operations to all parts where imbalances may arise.

To sum up, we analyze the reasons for the poor performance of AL in aerial images and propose targeted and complete solutions, and the contributions are as follows,
\begin{itemize}
\item According to the characteristics of aerial images, we propose an edge-guided and class-balanced AL method, which can make the network obtain competitive performance with only a few annotations. This method can be applied to any existing segmentation network. Furthermore, we set up a fair and strong benchmark for future research on AL for aerial image segmentation.
\item In segmentation tasks, the uncertain regions of aerial images are concentrated in the edge regions, so we propose an edge-guided labeling unit for this characteristic.
Using the edge regions as labeling units can include as many uncertain results in a unit, which can greatly reduce the number of manual labels while ensuring network training.
\item We analyze all stages and add class balance operations in all parts where imbalances may arise. An off-the-shelf pre-trained model is introduced to get the initial data efficiently.
Meanwhile, a performance-based balance is designed for the subsequent acquisition strategy, which shows an impressive effect. Besides, an adaptive threshold balance and balanced contrastive learning are set to keep the balance of pseudo labels and feature extraction, respectively.
\end{itemize}

\section{Related Work}

\subsection{Active Learning}
Active learning methods can be divided into two streams: uncertainty-based methods and diversity-based methods.

\emph{Uncertainty-based methods:} uncertainty has been widely used in active learning to estimate the importance of samples. It can be defined as the posterior probability of the predicted class \cite{acl19,acl18}, the posterior probability margin between the first and second predicted classes \cite{acl14}, or the entropy of the posterior probability distribution \cite{acl32}.
Yoo \cite{acl40} designed a module to learn how to predict unlabeled images' loss and select the ones with the highest predicted loss.
Gal \cite{acl8} obtained uncertainty estimates through multiple forward passes of Monte Carlo dropout.
Seung \cite{acl34} trained multiple models to construct a committee and measured uncertainty through consensus among multiple predictions of the committee. 

\emph{Diversity-based methods:} The purpose of diversity-based methods is to solve the problem of sampling bias in batch queries.
The core set method \cite{coreset} attempts to solve this problem by constructing a core subset. Context-aware methods \cite{acl10,acl23} take into account the distance between sample features and their surrounding points to enrich the diversity of labeled data sets. 

In active learning of semantic segmentation, uncertainty-based strategies are generally used.
Monte-Carlo dropout \cite{wk18} uncertainty is applied at the pixel level \cite{wk21,wk40}. In \cite{wk47}, the author experimented with five classification acquisition functions for segmentation tasks, including entropy-based, core-set \cite{coreset}, K-means, and Bayesian sampling \cite{wk18}, and applied them to electron microscope segmentation. In \cite{wk64}, the author combined two sampling items for histological image segmentation, one is uncertainty, and the other is diversity.

However, unlike the previous ones, our acquisition of initial labeled data is based on diversity, while the acquisition of the subsequent labeled data is based on uncertainty. Meanwhile, we explicitly add class balance into both acquisition strategies.

\subsection{Labeling Unit in Active Learning}
If the labeling unit is taken as the standard, active learning for semantic segmentation can be divided into image-level methods \cite{sp35,sp31,sp8} and region-level methods. The former treats the entire image as a labeling unit, while the latter divides the image into patches (regions) and treats each patch as a labeling unit. In the first category, Yang et al. \cite{sp35} used the model prediction and feature descriptors extracted from the trained CNN model to select a group of the most representative and uncertain samples for annotation. 
Currently, in the active learning of semantic segmentation, the mainstream method is to take regions (patches) as samples.
In the region-level methods, the method can be further divided into  regular region (such as rectangles) methods \cite{sp23,ralis,sp6} and  irregular region (such as superpixel) methods \cite{sp18,sp30}.
Grain \cite{sp23} fuses entropy and cost estimation to select information-rich but inexpensive annotations. RALIS \cite{ralis} uses reinforcement learning to learn the optimal strategy for regional selection. MetaBox+ \cite{sp6} selected areas based on predicted quality and cost estimates. 
SpAL \cite{spal} demonstrates that a superpixel-based approach can reduce annotation costs because it treats a superpixel block as a class.

We analyze the error-prone and uncertain regions in the segmentation of aerial images and propose a novel edge-guided labeling unit.
Using edge regions as labeling units can contain as many uncertain results as possible in one unit, which can significantly reduce the number of manual labels while ensuring network training.

\subsection{Class Balance in Active Learning}
Using class-imbalanced data for learning is a well-studied research issue \cite{ba32}.
At present, the commonly used methods for class balance can be divided into two categories: reweighting-based methods and resampling-based methods.
The network can reduce bias towards the most common categories by reweighting the samples in the training loss.
A popular method is to reweight a sample by the inverse of its class frequency \cite{ba30}.
Cui et al. \cite{ba12} improved this method and proposed to reweight samples with the actual effective number of its classes.
The other method is based on resampling \cite{ba28}, which keeps different sampling times for different samples in training. Ren et al. \cite{ba46} studied the training of unbalanced data combined with label noise.

Our work focuses on the imbalance that may occur during active learning cycles.
\cite{CBAL} indicates that class balance, as one of the goals of active learning, is crucial for unbalanced data sets.
Previous studies addressing class imbalance in active learning include \cite{ba54,ba1,ba7}. Among them, only \cite{ba1} has been applied to deep learning. However, it studies the order of active learning to ensure class balance, but it is an artificial method, so it is difficult to popularize on a large scale.
Recently, RALIS \cite{ralis} proposed a reinforcement learning method to consider class balance as one of the acquisition strategies, and SpAL \cite{spal} ensures that a superpixel block contains as few categories as possible to keep the data balance. CBAL \cite{CBAL} adds penalty terms to ensure class balance. However, RALIS and SpAL guarantee only partial balances. CBAL ensures the global balance of each class, but it aims at the image classification task. In active learning, the acquisition units of segmentation and classification are different. Classification is based on an image, while currently, the commonly used segmentation is based on regions.

We add class balancing operations wherever class imbalances may occur, discard the previous data-based balance and use a performance-based balance, through which the imbalance is significantly suppressed.

\begin{figure*}[htbp]
\centering
\includegraphics[scale=0.50]{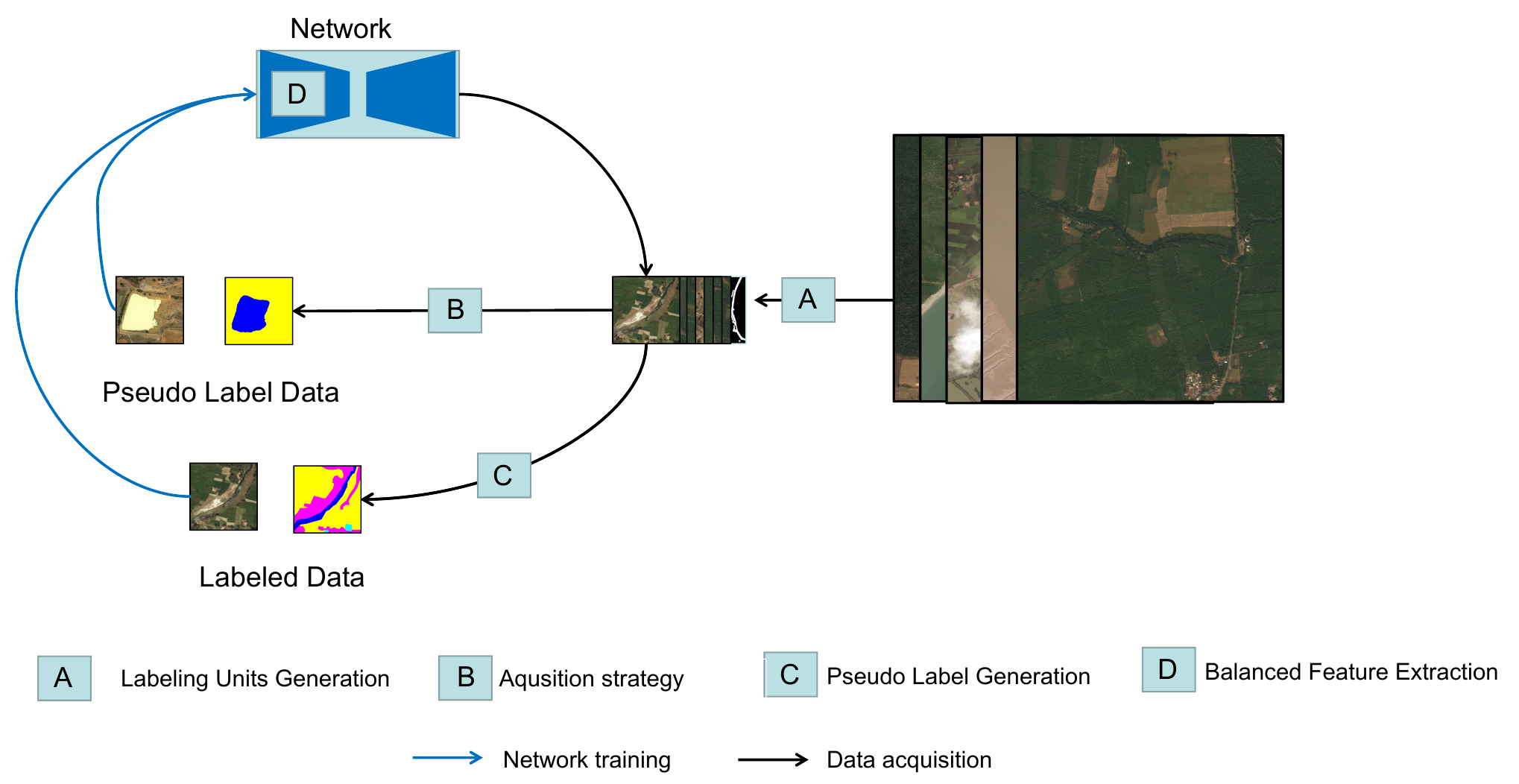}
\caption{
An overview of the overall process of active learning. First, the original data is divided into different labeling units, including rectangles and edge regions, and the process is shown in box A of the figure.
Then, the data to be labeled are selected through initial data acquisition, as shown in box C.
These labeled data are used for the initial training of segmentation networks.
The data to be labeled next is selected according to the initially trained network's output, which is also shown in box C.
For unselected data with high output confidence, pseudo-labels are generated from them to participate in network training, as shown in box B.
Box D represents the balanced contrastive learning in the feature extraction part of the segmentation network.
B, C, and D are all measures of class balance.
The blue lines represent network training, and the black lines represent data acquisition.
}
\label{overview}
\end{figure*}

\subsection{Self-Training and Supervised Contrastive Segmentation}
During self-training, the network generates pseudo labels and retrains the network based on these pseudo labels.
Wang et al. \cite{self50} design a traditional pseudo-label generating method and use the pseudo-label to train the existing semantic segmentation network to achieve better results than the previous traditional methods.
However, most of the works filter the network's output. \cite{incre} and \cite{selftrain} use the threshold strategy when selecting pseudo labels. Unlike these approaches, we set thresholds for each class and change them accordingly as the model changes. More importantly, we give more attention to the classes with poor performance.

Contrastive learning is used to model pixel relations in latent space \cite{contrastiveseg}.
It uses the loss function to pull the positive potential representation closer and the negative potential representation apart.
Besides, some methods \cite{hu23} propose to use the contrastive loss for unsupervised clustering of segmentation problems. However, the aforementioned works using contrastive learning focus on unsupervised learning. Without the guidance of labels, unsupervised contrastive learning may treat instances or pixels from different images as negative pairs, thus making errors.
Compared with unsupervised learning, supervised contrastive learning can enhance feature similarity within the same class without error and increase the ability to discriminate between different classes. \cite{contrastiveseg} is an outstanding work in supervised contrastive segmentation.
Our work employs supervised contrastive learning and
different from previous, our contrastive learning pays more attention to categories with poor performance.

\newpage
\section{Method}

\subsection{Overview}
The entire Active Learning (AL) process is shown in Fig. \ref{overview}. In the data processing part, the original unlabeled data is divided into different labeling units; we use a combination of rectangle and edge regions, as shown in A of Fig. \ref{overview}.
Then, the process enters AL iteration.
In the AL iteration, part of the data is first selected from the unlabeled data and manually labeled as the initial labeled data; next, the labeled data is used to train the network (Deeplabv3 \cite{deeplabv3} as default). Then, the data that needs to be labeled next is obtained according to the output of the trained network, i.e., the acquisition strategy, which is the process B shown in Fig. \ref{overview}. Meanwhile, the unlabeled part with high-confident outputs is selected as pseudo-labels to participate in network training, as shown in process C. The iterations continue until the network performance is satisfied or the amount of the labeled data reaches a budget (such as 10\% of the whole data). Throughout the process, people only need to label the data selected by the acquisition strategy.

Boxes A, B, C, and D in Fig.\ref{overview} represent the location of our contributions.
Among them, B, C, and D are used to ensure complete class balance.
Box A represents the generation of labeling units. We add edge-guided regions to the original rectangle regions, and the specific details are introduced in Section \ref{sec_lb_unit}.
Box C represents the acquisition strategy, including the balanced acquisition strategy for the labeled data at the initial and subsequent stages, which will be introduced in Section \ref{sec_balance_labeld}.
Box B represents a balanced pseudo-label generation for self-training. We use a way of setting different thresholds for different categories to make the generated pseudo-labels also balanced, which will be introduced in Section \ref{sec_balance_pseudo}.
Balanced contrastive learning in the network feature extraction part is introduced in Section \ref{sec_balance_contrastive}.

\subsection{Edge-guided Labeling Unit:}
\label{sec_lb_unit}
Errors in segmentation tasks often occur in the edge regions, and many previous works have also found this phenomenon, such as Dense$^{2}$Net \cite{denet} and \cite{edge_guided}.
Besides, the edge regions contain a large number of uncertain regions, as discussed in the Introduction section. Meanwhile, pseudo-labels are employed for self-training in areas with high confidence. Self-training and edge-guided labeling units are complementary to each other, i.e., self-training can obtain pseudo-label with high confidence inside the object according to the output of the network, but edge regions tend to be low confidence, so the labeling of edge regions is necessary.

\begin{figure}[ht]
\centering
\includegraphics[scale=0.22]{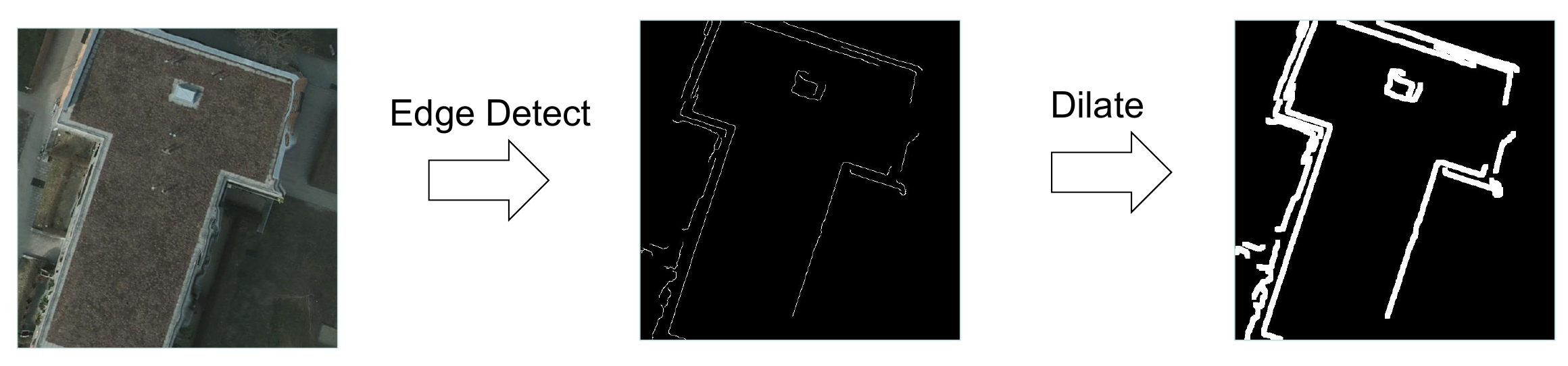}
\caption{
The process of obtaining an edge-guided labeling unit.
First, use the off-the-shelf edge detection algorithm to obtain the edge, and then use dilatation to obtain the edge regions.
The obtained edge regions are the labeling units.
}
\label{label_process}
\end{figure}

\begin{figure*}
\centering
\includegraphics[scale=0.80]{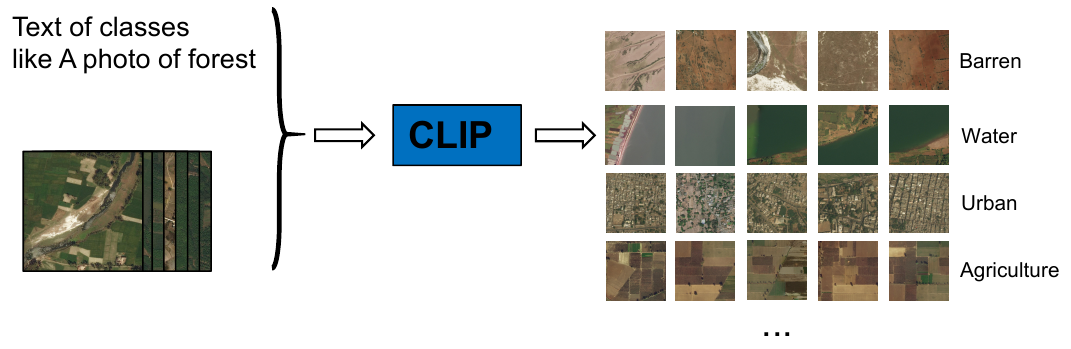}
\caption{
Overview of the initial data acquisition procedure.
The whole process is to put the regions of images into the CLIP to obtain the classes of the regions and then select the balanced samples according to the class of outputs.
}
\label{init_2}
\end{figure*}
Obtaining edge-guided labeling units is easy, as shown in Fig. \ref{label_process}.
We first use the edge detection algorithm in image processing to detect the edge and then dilate to get the edge regions.
Edge detection and dilation can be easily implemented through off-the-shelf codebases.
The edge detection algorithm can be the Canny detector or other network-based detectors.
We conduct experiments to verify the influence of edge-detection algorithms on the final performance.
Meanwhile, we found that by using a larger threshold at the beginning of training to select the most obvious edge area and then dynamically adjusting the threshold during subsequent training to select the less obvious edge area, the effect will be better.

There will also be edges inside some categories, such as the rectangular area of the building in Fig. \ref{label_process}. It is also necessary to label such regions.
Such regions are easily classified into other categories due to their different representations, so strong guidance is needed and necessary.
The edge-guided method only needs one mask (the mask of the edge regions). In contrast, the superpixel-based and region-based methods require M masks (M is the number of regions or superpixels), so edge-based labeling units are easy to implement.

\subsection{Balanced Acquisition Strategy of Labeled Data}
\label{sec_balance_labeld}
Previous methods \cite{deepact_weak,CBAL} generally use the same acquisition strategy for the initial and subsequent labeled data.
In fact, the initial labeled data has special significance for active learning, especially for the uncertainty-based strategy, which is critically relevant to the output of the model.
Intuitively, suppose the initially trained network is too biased and confident toward some classes with large sample sizes. In that case, the errors for small classes (with small sample sizes) will be consistent and even expanded. It is a vicious circle.
Based on this, we set different selection strategies for the initial and the subsequent labeled data acquisition strategies, which are introduced respectively in the following part.

\textbf{CLIP-based balance for the initially labeled data:}
The difficulty of the initial data acquisition is to make the selected samples include all classes, and each class is balanced without the help of the segmentation model.
Some works \cite{ralis} train initial models from in-game street scenes that can be automatically retrieved as the initial data, and other works use the pre-trained model from large-scale data sets like ImageNet \cite{imagenet}.
It can be concluded that the acquisition of initial data mainly uses existing models or data sets that can be easily obtained, which can be called prior knowledge.
We follow this idea to find suitable prior knowledge so that the classes of initial data are balanced.
At present, the core difficulty is that the existing available models are all aimed at natural images, and more seriously, these models are in poor generalization and thus work poorly in aerial images.

To solve this problem, we abandon the previous Core-set \cite{coreset} based methods and chose a method named Contrastive Language-Image Pre-training (CLIP), which employs texts rather than designed one-hot labels as the supervision information and has been demonstrated to contain strong generalization \cite{clip}.
The work based on CLIP has gained a lot of attention in the computer vision community and has achieved many important breakthroughs \cite{clip_methods}.

CLIP is a model trained with super-large image and text data. Due to the large scale of training data, CLIP can be extended to many non-seen classes, similar to zero-shot learning \cite{zero_shot}.
The process for selecting samples using CLIP is shown in Fig. \ref{init_2}.
We set the class name to the CLIP format as the text information and then the region as the image information.
The backbone used to extract image features is VIT-Large \cite{vit}. All networks are already trained and do not require finetuning.
And compared with core set-based methods, this method does not require clustering and only needs to be performed once to obtain all the initial data.

In this way, the CLIP model can output a class for each region, and then we select samples (regions) based on these outputs to make each class generally balanced.
Specifically, $N$ is the number of regions initially selected for manual annotation, and $C$ is the number of categories, so $\frac{N}{C}$ is the number of regions in each category.

\begin{figure*}
\centering
\includegraphics[scale=0.23]{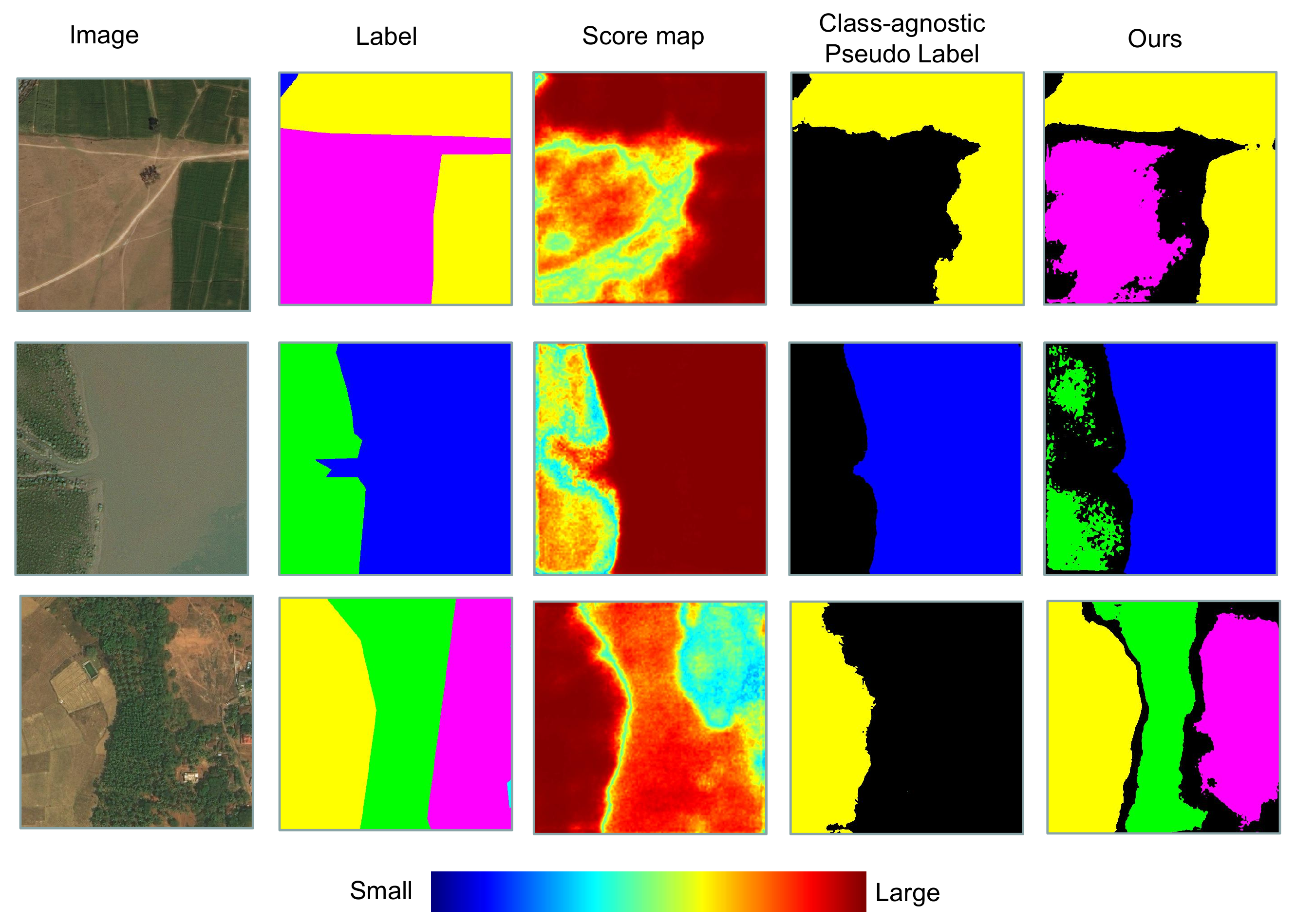}
\caption{
The display of the score map and different pseudo labels.
Black is the region where the score is below the threshold.
It can be seen that the score gap between different categories is very large. As shown in the second row, the score of the blue water is significantly larger than that of the green forest.
Class-agnostic approaches use the same threshold so that certain classes with lower scores are completely ignored.
Our method can set different thresholds for various classes and pay more attention to classes with poor performance, leading to very comprehensive pseudo labels.
}
\label{pic_pseudo_label}
\end{figure*}

\textbf{Performance-based balance for the subsequent acquisition of labeled data:}
Unlike the initial acquisition strategy, we have the initially trained segmentation network in the subsequent stage.
Therefore, we can choose which samples need to be labeled according to the network output.
Our main idea is to balance classes according to the performance of each class, which is intuitive and effective.

Firstly, we get the information of the region, i.e., the sum of the information (like uncertainty) of each pixel belonging to the region.
The pixel-level information in our work is measured by entropy.
We use $\mathbf{y}$ to denote the model's output of a region, and $y_{i, j, c}$ denotes the model's output of position $i$,$j$ for class $c$.
And there are total $C$ classes, and the region is $H \times W$.
Thus, the information $\mathcal{U}$ in the region is,
\begin{equation}
\label{information}
\mathcal{U}=-\sum_{i}^{H}\sum_{j}^{W}\sum_{c=1}^{C} y_{i, j, c} \log (y_{i, j, c})
\end{equation}
For each pixel in the region, we use the output $\mathbf{y}$ of the network through the argmax operation as the pseudo-class label $\mathbf{y}^{*}$, i.e
\begin{equation}
\mathbf{y}^{*}=\arg \max (\mathbf{y})
\end{equation}
where $\mathbf{y} \in \mathbb{R}^{H \times W \times C}$ and $\mathbf{y}^{*} \in \mathbb{R}^{H \times W}$. $\mathbf{y}^{*}$ is important to performance-based balance.

According to the pseudo-label map $\mathbf{y}^{*}$, we can get the proportion $N_c$ of pixels belonging to class $c$ in the region, i.e.,
\begin{equation}
N_c=\frac{1}{H \times W} \sum\limits_{i}^{H} \sum\limits_{j}^{W} \mathbbm{1}\left[y_{i,j}^{*}=c\right] ,
\end{equation}
where $\mathbbm{1}\left[{y}_{i,j}^{*}=c\right]$ indicates the pixels in location i,j of pseudo label associated to $c$, and when the pseudo label corresponding to the pixel is class $c$, the value is 1; otherwise, the value is 0.

Meanwhile, we can obtain the performance (such as Intersection over Union (IoU)) of the model for each class on the labeled data, and the calculated performance can guide us to choose which class needs to be sampled greatly.
Intuitively, for a class, if its IoU is low, it needs more samples. Based on the above analysis, we use $p_c$ to denote the IoU of class $c$ in the labeled data after normalization so that the sum is 1.
Therefore, the score obtained based on performance $\mathcal{P}$ is,
\begin{equation}
    \mathcal{P}= \sum_{c}^{C} \frac{1}{p_c} N_c.
\end{equation}
This actually gives a higher weight to the classes with poor performance, but the score is calculated in terms of classes rather than samples.

By combining the balance and the information amount, the comprehensive score $\mathcal{S}$ of the region is,
\begin{equation}
    \mathcal{S}= \mathcal{U} \times sigmoid (\mathcal{P}),
\end{equation}
and the score of other regions is calculated in the same way.
Since balance plays a moderating role, we add the sigmoid function in front of $\mathcal{P}$ to avoid it dominating the score.
Finally, we sort all the regions according to the final comprehensive scores and select the ones with the high scores to label them until the number of regions reaches a budget.


\subsection{Class-specific and Adaptive Balance for Pseudo Labels of Self-Training}
\label{sec_balance_pseudo}
Setting pseudo-labels for unlabeled regions is an important way to self-training \cite{self13,self14}.
The key to self-training is how to select the highly confident outputs as pseudo labels.
Most work uses thresholds to select low entropy or high score values, and the predicted pseudo labels tend to be overconfident, which may mislead network training and thus impair learning performance \cite{self53}.
Besides, the previous methods employ fixed thresholds, or class-agnostic adaptive thresholds \cite{incre}, which is very unfriendly to classes with poor performance.

Therefore, we set different adaptive thresholds for various classes, i.e., the class-specific and adaptive threshold balance.
Fig. \ref{pic_pseudo_label} shows the comparison between our and previous methods. It can be seen that the score gap between each category is very large. The previous method directly ignores some categories with low scores, and our method can avoid this problem by class-specific thresholds.
More importantly, we set a lower threshold for classes with poor performance to obtain more pseudo-labels.


Specifically, when the training of the network under the current labeled data set is completed, and the unlabeled data is input to the network for inference, the corresponding class and score of each pixel are obtained.
The first $K_c$ outputs with the high score are then selected as the pseudo label for class $c$.
$K_c$ is the ratio threshold of class c, which is related to the performance of class $c$, i.e.,
\begin{equation}
    K_c=base \times e^{\bar{\mathcal{P}} -p_c},
\end{equation}
where $base$ represents the basic selection ratio, which is set $1/2$ as default, and $\bar{\mathcal{P}}$ represents the mean of performance (like mIoU) of all classes. $p_c$ denotes the performance of class $c$.
If class $c$ gets poor performance, i.e., $p_c$ is small, then $K_c$ is larger, and so class $c$ can get a higher percentage of pseudo labels. 
In this way, the poorly performing classes can get more pseudo-labels.
If $K_c$ is greater than 1, then we take 1. In practice, the maximum $K_c$ is about 0.9.
It should be noted that the pixels that are not labeled and not selected as pseudo labels do not participate in network training.

\subsection{Balanced Contrastive Learning for Feature Extraction}
\label{sec_balance_contrastive}
Both the above acquisition of labeled data and generation of pseudo labels are carried out simultaneously on the feature extractor and classifier. And \cite{decouple_im} proposes that decoupling feature representation and classification has a good effect on the imbalance.
Besides, we should fully use the labeled data of the classes with poor performance, and the edge regions often contain multi classes, which is suitable for contrastive learning.
Therefore, we conduct contrastive learning according to the labeled image parts for the feature extraction part.
More importantly, we conduct a balance to contrastive learning.

Specifically, we use the features of the classes with poor performance as anchors to calculate the contrastive loss.
The class with poor performance is defined as: if the IoU of this class is lower than the mean IoU, it is a poor-performing class.
We use $\boldsymbol{i}$ to represent the anchor feature selected from the classes with poor performance, and $\boldsymbol{i}^{+}$ is the feature of the same class, so $\boldsymbol{i}^{+}$ can be regarded as the positive sample for $\boldsymbol{i}$, and $\boldsymbol{i}^{-}$ is the feature from different classes and thus is the negative sample. The downsampled label determines the class of each feature. Therefore, the supervised contrastive loss $\mathcal{L}_{nce,i}$ for anchor feature $\boldsymbol{i}$ is,
\begin{small}
\begin{equation}
\mathcal{L}_{nce,i}=\frac{1}{|\mathcal{P}_{i}|} \sum_{\boldsymbol{i}^{+} \in \mathcal{P}_{i}}-\log\frac{\exp \left(\boldsymbol{i} \cdot \boldsymbol{i}^{+} / \tau\right)}{\exp \left(\boldsymbol{i} \cdot \boldsymbol{i}^{+} / \tau\right)+\sum\limits_{i^{-} \in \mathcal{N}_{i}} \exp \left(\boldsymbol{i} \cdot \boldsymbol{i}^{-} / \tau\right)},
\end{equation}
\end{small}
where $\mathcal{P}_{i}$ denotes the set of positive samples of $\boldsymbol{i}$, and $\mathcal{N}_{i}$ is negative. $|\mathcal{P}_{i}|$ denotes the number of positive samples. '$\cdot$' denotes the inner (dot) product, and $\tau>0$ is a temperature hyper-parameter, and we set it as 0.1 for default. Note that all the features in the loss function are $\ell_{2}$-normalized.
The idea of the formula is to make the poorly performing classes better separated from the rest.

\section{Experiments}
In this section, extensive experiments are conducted to verify the effectiveness of our method.
Implementation details are introduced in section \ref{imple_detail_sec}, including data sets used, experimental parameters, and the evaluation metric.
Section \ref{sec_sota} is the comparison with other state-of-the-art (SOTA) methods. The comparison with the baseline and other different acquisition strategies is shown in Section \ref{sec_baseline}.
Section \ref{sec_data_analysis} shows the class proportion of labeled data.
The results with different segmentation networks are shown in Section \ref{sec_segmentation_network}
In section \ref{ablation_sec}, we conduct sufficient ablation experiments to verify the roles of edge-guided labeling units, CLIP-based initial balance, performance balance, pseudo-label balance, and feature extraction balance. The effects of various hyper-parameters on the final results are introduced in Section \ref{sec_param}.

\subsection{Implementation Details}
\label{imple_detail_sec}
\textbf{Dataset:}We use Deepglobe \cite{deepglobe}, Potsdam, and Vaihingen data sets.
Deepglobe consists of 803 images, each with a resolution of $2,448, \times 2,448$. In line with \cite{glnet}, we divide these images into training and test sets of 455 and 348, respectively. The dense pixel-level label is divided into six classes: urban, agricultural, rangeland, forest, water, and barren.
We use cross-validation, i.e., 20\% of the training set was used for the validation set.
"unknown" region is not considered in the challenge as previous methods \cite{glnet,incre}. 
We crop an image to patches with the size of $800 \times 800$ sequentially so that the training set has 4095 images.

The Vaihingen dataset is composed of 33 aerial images collected of the city Vaihingen. The average size of each image is 2494 $\times$ 2064 pixels, Among these images, 16 of them are manually annotated with pixel-wise labels, and each pixel is classified into one of six land cover classes (buildings, cars, low vegetation, impervious surfaces, trees, and clutter/background). Following the setup in \cite{tgrs_ccanet}, we select 11 images for training, and the remaining five images (image IDs: 11, 15, 28, 30, 34) are used to test our model.
 
The Potsdam dataset consists of 38 high-resolution aerial images.
The size of all images is 6000 × 6000 pixels, annotated with pixel-level labels of six classes as the Vaihingen dataset. To train and evaluate networks, we utilize 10 images for training and build the test set with the remaining images (image IDs: 02 11, 02 12, 04 10, 05 11, 06 07, 07 08, 07 10), which follows the setup in  \cite{tgrs_ccanet}.

The proportions of categories are shown in Table \ref{table_classes_deepglobe_pots_vaihin}, and it can be seen that they are very unbalanced.

\begin{table*}[htbp]
\centering
  \caption{Propotion for each class on Deepglobe, Vaihingen, and Potsdam.}
  \label{table_classes_deepglobe_pots_vaihin}
  \scalebox{0.92}{
  \begin{threeparttable}
  \begin{tabular}{c|c|c|c|c|c|c|c}
    \toprule
\multirow{2}{*}{Deepglobe}&Class&Urban &Agriculture&Rangeland&Forest&Water&Barren\\
\cmidrule{2-8}
&Proportion&9.3&\textcolor{blue}{57.7}&10.2&13.8&\textcolor{red}{3.7}&6.1\\
\midrule
\multirow{2}{*}{Vaihingen}&Class&Impervious Surfaces&Buildings&Low Vegetation&Trees&Cars&Cluster\\
\cmidrule{2-8}
&Proportion&\textcolor{blue}{30.8}&28.8&20.3&13.7&\textcolor{red}{1.8}&4.5\\
\midrule
\multirow{2}{*}{Potsdam}&Class&Impervious Surfaces&Buildings&Low Vegetation&Trees&Cars&Cluster\\
\cmidrule{2-8}
&Proportion&\textcolor{blue}{29.0}&26.7&20.0&22.0&1.2&\textcolor{red}{0.9}\\
\bottomrule
\end{tabular}
\begin{tablenotes} 
		\item \textcolor{blue}{Blue} represents the proportion of the largest class and \textcolor{red}{red} represents the proportion of the smallest class.
     \end{tablenotes} 
\end{threeparttable}
}
\end{table*}

\textbf{Implementation details:}
Unless otherwise noted, all the methods mentioned in this section use Deeplabv3 \cite{deeplabv3} as the base segment network, with ResNet-50 \cite{resnet} as its backbone, and are initialized with a pre-trained model on ImageNet \cite{imagenet}.
We also use other segmentation networks to verify the validity of our method.
The segmentation loss of all methods is the cross-entropy loss as previous \cite{danet,deeplabv3}.
We train the network using SGD as the optimizer.
The initial learning rate (lr) is $10^{-2}$. Gamma is 0.998 (lr = initial lr $\times$  ($Gamma^{epochs}$)). Weight-decay is $1e-4$ and momentum is 0.95.
Since the GPU memory requirements are a key issue for semantic segmentation, in-place batch normalization is used, as proposed in \cite{bn}.
We use image flipping as the data augmentation.
During training and testing, images are cropped to $800 \times 800$.
If the size of a region is not specified, it is $80 \times 80$ as the default. In this case, an image has 100 regions, and each region does not overlap with the other.
For each iteration, we randomly select 100 images (10,000 regions) from all the unlabeled data and then select 500 regions according to our acquisition strategy. After the acquisition, the model trained 50 epochs. Do this until the annotated data reach a budget.
If there is no special state, 5\% are selected from all the unlabeled data as the initial labeled data.

In balanced contrastive learning, for the classes with poor performance, we select the features of $T$ pixel as anchor features.
For each anchor feature, 512 and 1024 pixel features are selected as positive samples and negative samples, i.e., $\mathcal{P}_{i}$=512 and $\mathcal{N}_{i}$=1024.
$T$ is a hyperparameter, and we have done several experiments to verify the influence of $T$ on the final result.
It should be noted that the memory bank is not added to our work, so it is a batch internal comparison, which simplifies the training process.

Before edge detection, a Gaussian filter with a kernel size of $17 \times 17$ is used for smoothing.
Unless otherwise specified, we use the Canny edge detection method to detect edges, and the maximum and minimum thresholds are 10 and 80.
The kernel size of dilation is $9 \times 9$.
The threshold is updated gradually, i.e., every time the budget increases by 5\% of the total data volume, and the threshold is reduced by 5.

\begin{figure*}[htbp]
\centering
\includegraphics[scale=0.45]{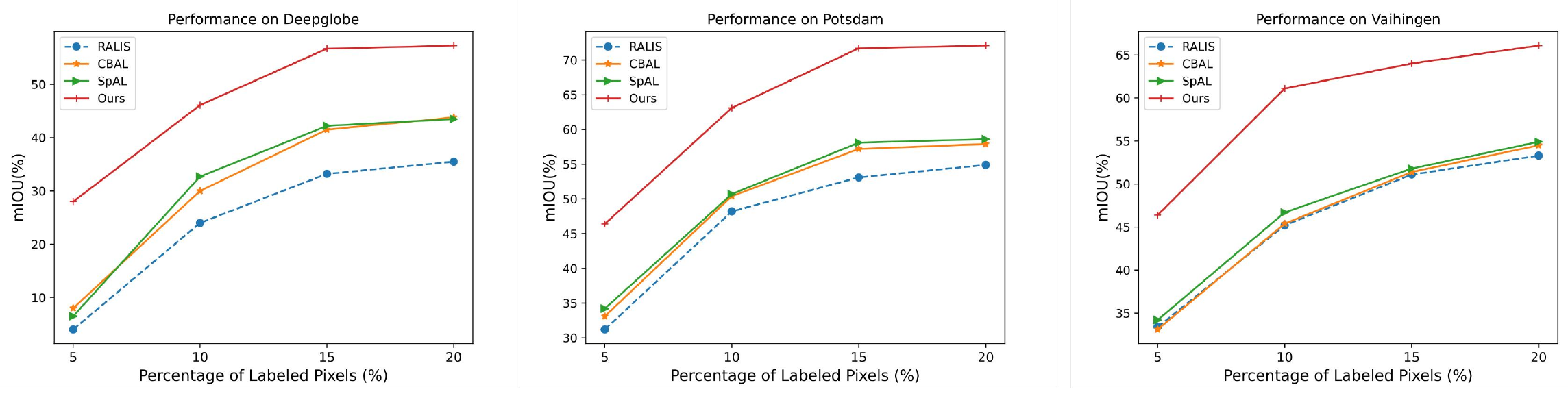}
\caption{
The comparison with other SOTA methods on Deepglobe, Potsdam, and Vaihingen datasets.
The figure shows the results where the labeled data accounts for 5\%, 10\%, 15\%, and 20\% of the entire data.
}
\label{pic_result_sota}
\end{figure*}

\begin{figure*}[htbp]
\centering
\includegraphics[scale=0.45]{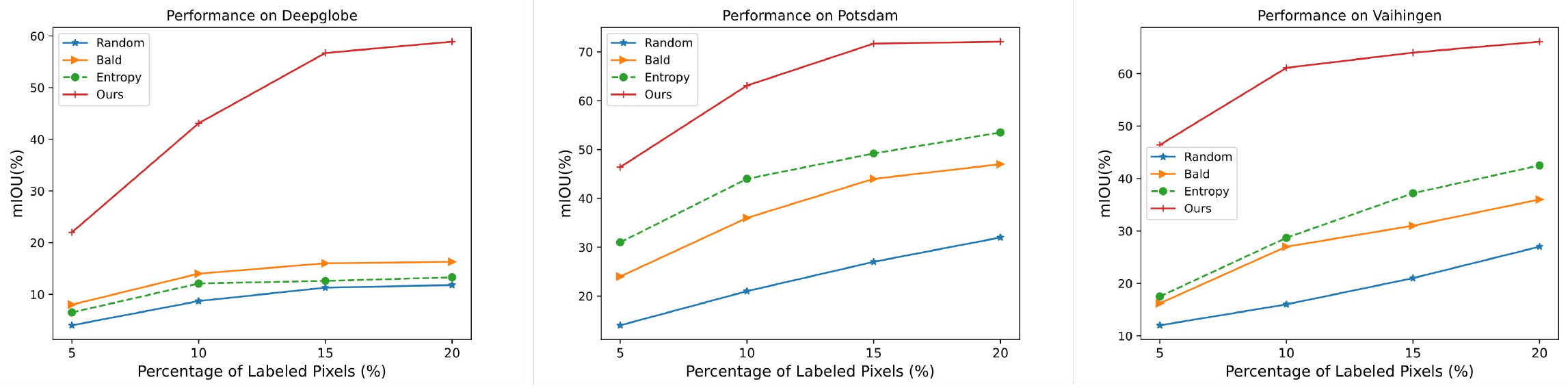}
\caption{
The comparison with other acquisition strategies and the baseline on Deepglobe, Potsdam, and Vaihingen datasets. The entropy method is the baseline.
The figure shows the results where the labeled data accounts for 5\%, 10\%, 15\%, and 20\% of the entire data.
}
\label{pic_result_others}
\end{figure*}

\textbf{Evaluation metric:}
We use mean Intersection over Union (mIoU) as the evaluation metrics.
mIoU is the ratio of the intersection and union of two sets of true and predicted values and can be evaluated by
\begin{equation}
\mathrm{mIoU}=\frac{1}{K} \sum_{i=1}^{K} \frac{p_{i i}}{\sum_{j=1}^{K} p_{i j}+\sum_{j=1}^{K} p_{ji}-p_{i i}},
\end{equation}
where $K$ is the number of classes, $i$ is the true class label, and $j$ is the predicted label.
$p_{i j}$ denotes the number of pixels of class $i$ predicted to class $j$.

$\mathrm{mean\ F}1$ score is defined as
$$
\mathrm{mean\ F}1=\left(1+\beta^2\right) \frac{\text { precision }}{\beta^2 \text { precision }+\text { recall }},
$$
where $\beta$ is the equivalent factor between precision and recall,  and $ \beta$ is set as 1. In addition, precision and recall are defined as
$$
\begin{aligned}
\text { precision } &=\frac{\mathrm{TP}}{\mathrm{TP}+\mathrm{FP}} \\
\text { recall } &=\frac{\mathrm{TP}}{\mathrm{TP}+\mathrm{FN}} .
\end{aligned}
$$

\begin{table*}[htbp]
\centering
  \caption{class-level results on deepglobe}
  \label{results_classes_deepglobe}
  \scalebox{0.9}{
    \begin{threeparttable}
  \begin{tabular}{c|c|c|c|c|c|c|c}
    \toprule
Method     &Urban &Agriculture&Rangeland&Forest&Water&Barren&mIoU (\%)\\
\midrule
Random&3.1&54.4&2.5&4.5&4.3&2.1&11.8\\
Entropy (baseline)&6.1&72.1&4.8&5.8&5.1&3.9&16.3\\
BALD\cite{bald3}&4.2&60.9&3.9&5.9&2.4&2.7&13.3\\
\midrule
RALIS \cite{ralis}&32.1&65.2&17.9&39.5&39.2&18.9&35.5\\
SpAL \cite{spal}&\textcolor{blue}{49.9}&\textcolor{blue}{72.4}&25.5&38.4&41.1&\textcolor{blue}{35.3}&\textcolor{blue}{43.8}\\
CBAL \cite{CBAL}&43.6&72.2&\textcolor{blue}{26.8}&\textcolor{blue}{44.0}&\textcolor{blue}{43.2}&30.9&43.5\\
Ours&\textbf{69.2}&\textbf{79.1}&\textbf{46.9}&\textbf{68.0}&\textbf{57.4} &\textbf{52.1}&\textbf{62.1} (+13.8)\\
\bottomrule
\end{tabular}
\begin{tablenotes} 
		\item Blue represents the second-best result, and values in parentheses are the difference between our method and the second-best result.
     \end{tablenotes} 
\end{threeparttable}
}
\end{table*}

\begin{table*}[htbp]
\centering
  \caption{class-level results on Potsdam}
  \label{results_classes_potsdam}
  \scalebox{0.9}{
    \begin{threeparttable}
  \begin{tabular}{c|c|c|c|c|c|c|c}
    \toprule
Method&Impervious Surface&Building&Low Vegetation&Tree&Car&$\mathrm{mean\ F}1$ (\%)&mIoU (\%)\\
\midrule
Random&83.1&82.2&56.7&51.1&38.8&62.4&32.3\\
Entropy (baseline)&83.1&81.7&72.2&69.0&63.2&73.8&53.5\\
BALD\cite{bald3}&72.1&74.5&72.1&59.3&60.1&67.6&47.7\\
\midrule
RALIS \cite{ralis}&80.1&\textcolor{blue}{84.2}&71.1&67.0&69.7&74.4&54.9\\
SpAL \cite{spal}&\textcolor{blue}{83.5}&77.9&\textcolor{blue}{74.7}&72.1&\textcolor{blue}{77.9}&\textcolor{blue}{77.2}&\textcolor{blue}{58.6}\\
CBAL \cite{CBAL}&81.2&78.9&72.2&\textcolor{blue}{71.3}&77.2&76.2&57.9\\
Ours&\textbf{90.4}&\textbf{92.0}&\textbf{84.1}&\textbf{85.3}&\textbf{92.1}&\textbf{88.8} (+11.6)&\textbf{72.1} (+13.5)\\
\bottomrule
\end{tabular}
\begin{tablenotes} 
		\item Blue represents the second-best result, and values in parentheses are the difference between our method and the second-best result.
     \end{tablenotes} 
\end{threeparttable}
}
\end{table*}

\begin{table*}[htbp]
\centering
  \caption{class-level results on Vaihingen}
  \label{results_classes_vaihingen}
  \scalebox{0.9}{
    \begin{threeparttable}
  \begin{tabular}{c|c|c|c|c|c|c|c}
    \toprule
Method&Impervious Surface&Building&Low Vegetation&Tree&Car&$\mathrm{mean\ F}1$ (\%)&mIoU (\%)\\
\midrule
Random&79.2&85.5&42.3&33.5&27.9&53.7&27.0\\
Entropy (baseline)&78.8&84.6&60.9&61.1&43.3&65.7&42.5\\
BALD\cite{bald3}&74.2&84.3&55.1&55.6&41.2&62.0&36.9\\
\midrule
RALIS \cite{ralis}&78.2&89.1&\textcolor{blue}{69.6}&74.2&53.1&72.8&53.3\\
SpAL \cite{spal}&\textcolor{blue}{82.4}&85.9&68.4&78.2&\textcolor{blue}{58.1}&\textcolor{blue}{74.6}&\textcolor{blue}{54.9}\\
CBAL \cite{CBAL}&81.2&\textcolor{blue}{89.2}&67.3&\textcolor{blue}{79.8}&55.2&74.5&54.5\\
Ours&\textbf{88.7}&\textbf{91.1}&\textbf{74.2}&\textbf{87.9}&\textbf{90.2}&\textbf{86.4} (+11.8)&\textbf{66.1} (+11.2)\\
\bottomrule
\end{tabular}
\begin{tablenotes} 
		\item Blue represents the second-best result, and values in parentheses are the difference between our method and the second-best result.
     \end{tablenotes} 
\end{threeparttable}
}
\end{table*}

\subsection{Comparison with SOTA methods}
\label{sec_sota}
We first introduce how the benchmark is established and then compare our method with others.

\textbf{Benchmark building:}
Previous active learning methods focuse on class balancing, include inserting class balancing in optimization goals and making balanced acquisition strategies.

\emph{RALIS} \cite{ralis} applies reinforcement learning to active learning and strives to maximize the IoU of each class, so compared with other methods, it indirectly requires more labels for the areas where the class is underrepresented.

\emph{SpAL} \cite{spal} uses superpixels instead of rectangular boxes for the label, and it proposes a class-balanced acquisition function that further improves the performance of the super pixel-based approach by favoring samples from poor-performing object classes, i.e., selecting a pseudo-label containing as many categories of the superpixel blocks as possible.

\emph{CBAL} \cite{CBAL} puts forward a general optimization framework that explicitly considers class balance and directly puts class balance into the optimization objective for image classification. We changed the image- level to the region-level so that it can be used for image segmentation.

In order to ensure fairness, we choose the best parameters of these methods, i.e., the parameters used are the same as the original paper.
After the above operations, we have set up a fair benchmark for the balance of active learning methods of aerial image semantic segmentation.

\textbf{Comparison with SOTA methods:}
We first show the results of the method under different budgets and then introduce the performance of each category under the same budget.
The results compared with other state-of-the-art (SOTA) methods under various budgets are shown in Fig. \ref{pic_result_sota}.
The performance of CBAL and SpAL is similar, and both are better than RALIS, indicating that direct balancing is more effective.
Our method outperforms other methods by rank.
For different data sets, the various methods show the highest differentiation on Deepglobe, followed by Potsdam and Vaihingen. Deepglobe is the biggest and most challenging dataset. Potsdam and Vaihinge have smaller data sizes.
The size and difficulty of datasets may be the reasons for the different performance of the methods on different datasets.

The goal of our work is to make those classes with small sample sizes and poor performance achieves good performance, so class-level analysis is necessary.
The class-level performance with 20\% budget is shown in Table \ref{results_classes_deepglobe}, Table \ref{results_classes_potsdam}, and Table \ref{results_classes_vaihingen}.
Taking Deepglobe as an example, by analyzing the performance of various methods, it can be seen that methods can be divided into three levels.
The first level is the previous methods that do not pay attention to class balance, and their effects are very poor. 
These methods include Random, Entropy, and BALD.
The IoU of most classes is less than 10\%, while the best class agriculture reaches 70\%, so the performance between classes is seriously imbalanced.
And the mIoU is less than 20\%.
The second level is the methods that pay attention to class balance.
These methods include RALIS, SpAL, and CBAL.
Compared with the methods that pay no attention to balance, these methods achieve significant improvements, but there is still a big gap compared with ours.
Our method improves the effect of class balance from various aspects.
Among them, class water, forest, barren, and urban are all greatly promoted. These comparisons fully demonstrate the effectiveness of our work.
We used F1 as another evaluation metric on the Potsdam and Vaihingen datasets, and it can be seen that the obtained results are very similar to the IoU in Deepglobe.

\subsection{Comparison with Baselines and Other Acquisition Strategies}
\label{sec_baseline}
\textbf{Introduction of other methods:}
For the calculation of the amount of information (Eq. \ref{information}), besides the entropy method, we also use other methods such as random, Bayesian Active Learning by Disagreement (BALD) \cite{bald1}, to verify the validity of our class balanced strategy.

\emph{Random}: In each training batch, regions are randomly selected and incorporated into the labeled set.

\emph{Entropy}: It purely aims to select those regions with the least confidence predicted by the network. The entropy of a region is computed as the average entropy of all pixels within that region.

\emph{Bayesian Active Learning by Disagreement (BALD)} \cite{bald1,bald2,bald3}: Shannon entropy is used to measure the amount of information in the sample (region).
The difference in information entropy represents the difference between the average information entropy of a sample and that of the whole. The larger the difference is, the more information the sample contains.
BatchBALD \cite{bald3} is used for the implementation as previous work \cite{ralis}.

\textbf{Performance analysis:}
We also conduct analyses of different budgets and analyses at the class level.
The experimental results under various budgets are shown in Fig. \ref{pic_result_others}.
The method entropy is the baseline.
It can be seen that class balance plays a significant role in all situations and all datasets.
This shows that our method of class balance can effectively select the most valuable samples for each class. 
The class-level analysis is shown in Table \ref{results_classes_deepglobe}, Table \ref{results_classes_potsdam}, and Table \ref{results_classes_vaihingen}, showing very severe class imbalance, which also demonstrates the practical value of our work.

\subsection{Labeled Data Analysis:}
\label{sec_data_analysis}
\begin{table}[htbp]
\centering
  \caption{Proportion for each class of different methods on deepglobe}
  \label{results_label_classes}
  \scalebox{0.85}{
  \begin{tabular}{c|c|c|c|c|c|c}
    \toprule
Method     &Urban &Agri.&Range.&Forest&Water&Barren\\
\midrule
Random&8&61&10&14&2&5\\
Entropy&6&72&5&9&3&5\\
BALD\cite{bald3}&10&52&7&19&3&10\\
\midrule
RALIS \cite{ralis}&12&56&12&8&3&9\\
SpAL \cite{spal}&14&51&14&7&4&10\\
CBAL \cite{CBAL}&9&51&10&11&6&13\\
Ours&15&40&14&13&5&13\\
\bottomrule
\end{tabular}
}
\end{table}

The proportion of different classes on labeled data by various methods is analyzed, as shown in Table \ref{results_label_classes}. We analyze the situation of each class when the annotation is the largest, i.e., the budget is 20\%.
It can be seen that even if we do not use data balance, the data obtained by our method is relatively balanced.
As expected, we have a relatively large sample size of categories for some indistinguishable categories, such as urban. Compared with other methods, our method can get relatively more balanced labeled data.

\subsection{Results on different segmentation networks:}
\label{sec_segmentation_network}
\begin{table}
\centering
  \caption{Results on Deepglobe with different segmentation networks}
  \label{table_result_deepglobe}
    \scalebox{0.84}{
   \begin{tabular}{c|c|c}
    \toprule
Segmentation Network&Active Learning Method&mIoU (\%)\\
\midrule
\multirow{4}{*}{PSPNet \cite{pspnet}}&RALIS \cite{ralis}&14.2\\
&SpAL \cite{spal}&\textcolor{blue}{41.0}\\
&CBAL\cite{CBAL}&37.1\\
&Ours&\textbf{56.3}\\
\cmidrule{2-3}
&All Data&58.1\\
\midrule
\multirow{4}{*}{DANet \cite{danet}}&RALIS\cite{ralis}&19.4\\
&SpAL\cite{spal}&\textcolor{blue}{42.1}\\
&CBAL\cite{CBAL}&31.2\\
&Ours&\textbf{57.6}\\
\cmidrule{2-3}
&All Data&58.4\\
\midrule
\multirow{4}{*}{HRNet+OCR \cite{ocr}}&RALIS\cite{ralis}&38.6\\
&SpAL\cite{spal}&\textcolor{blue}{50.4}\\
&CBAL\cite{CBAL}&45.7\\
&Ours&\textbf{65.1}\\
\cmidrule{2-3}
&All Data&68.9\\
\midrule
\multirow{4}{*}{S-RA-FCN \cite{s-ra-fcn}}&RALIS\cite{ralis}&40.1\\
&SpAL\cite{spal}&\textcolor{blue}{51.2}\\
&CBAL\cite{CBAL}&46.1\\
&Ours&\textbf{65.8}\\
\cmidrule{2-3}
&All Data&69.7\\
  \bottomrule
\end{tabular}
}
\end{table}

To verify the effectiveness and generalizability of the proposed method, we also carried out experiments on PSPNet \cite{pspnet}, DANet \cite{danet}, HRNet+OCR \cite{ocr}, and S-RA-FCN \cite{s-ra-fcn} in addition to Deeplabv3 \cite{deeplabv3}.
The experiments are conducted on the Deepglobe dataset, and the proportion of labeled data is 20\% of the total.
The experimental results are shown in Table \ref{table_result_deepglobe}.
All the experimental settings are the same as before, except for the segmentation network.
It can be seen that the performance of the active learning methods is related to the segmentation network.
Our methods, SpAL, and RALIS, perform better on DANet than PSPNet, while CBAL performs better on PSPNet than DANet.
However, regardless of the network, our method achieves the best results and is far ahead of other methods, which fully demonstrates the effectiveness and generality of our method.

\subsection{Ablation Study}
\label{sec_ablation}
We conduct ablation studies in five aspects: edge-guided labeling units, initial data acquisition balance, subsequent acquisition balance, pseudo-label balance, and contrastive learning for feature extraction balance. Therefore, the respective effects of these five aspects will be verified in this section. 
Due to the large data size and the obvious distinguishment for each method, ablation experiments are conducted on the Deepglobe dataset. 
We use Deeplabv3 as the segmentation network, and the labeled data stations account for 20\% of the data.

\begin{table}[htbp]
\centering
  \caption{the effect of edge-guided labeling units}
  \label{table_edge}
    \scalebox{1.0}{
\begin{tabular}{c|c}
\toprule
edge-guided labeling units&mIoU (\%)\\
\midrule 
&58.9\\
\Checkmark&62.1\\

\bottomrule
\end{tabular}
}
\end{table}

\begin{figure}[htbp]
\centering
\includegraphics[scale=0.22]{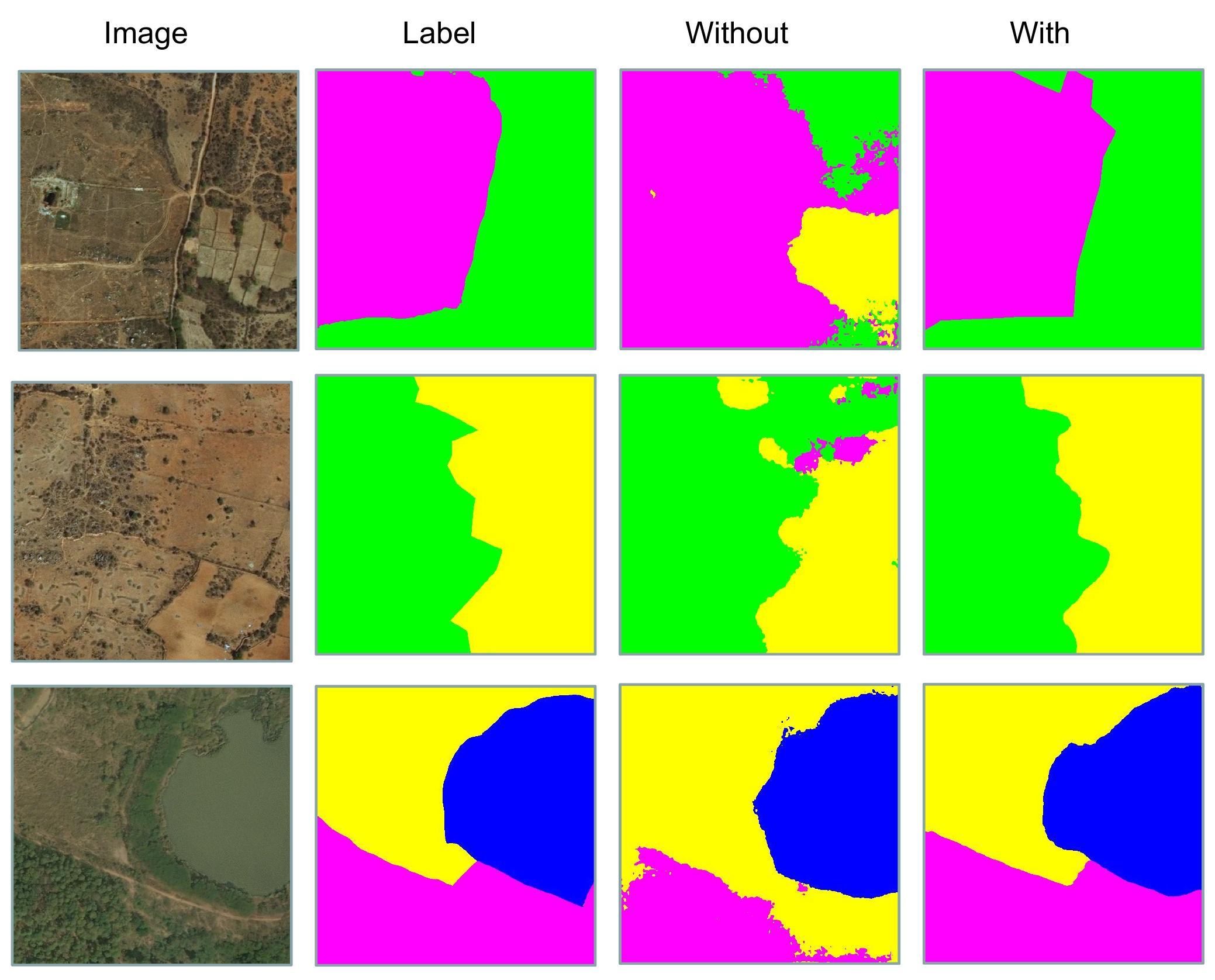}
\caption{
Visualization results with and without edge-guided labeling units on the Deepglobe dataset.
}
\label{pic_edge_label}
\end{figure}

\textbf{The effect of edge-guided labeling units:}
To verify the influence of edge parts on the final results, we conduct experiments with and without edge-guided labeling units in Deepglobe. The results are shown in Table \ref{table_edge}.
The visualization results are shown in Fig. \ref{pic_edge_label}.
It can be seen that the edge-guided labeling units not only make the edges better but also make the categories more discriminative.
We use contrastive learning in the feature extraction part of the network, and contrastive learning requires an image or a batch to contain as many categories as possible.
Multiple categories can be obtained at one time by labeling edge regions to improve the effect of contrastive learning. Therefore, edge-guided labeling units are of great significance, not limited to the performance improvement of edge regions. In general, edge-guided labeling units are necessary.

\label{ablation_sec}

\begin{table}[htbp]
\centering
  \caption{the effect of different initial balances}
  \label{result_initial}
    \scalebox{1.0}{
\begin{tabular}{c|c|c|c}
\toprule
Random &Coreset &CLIP&mIoU (\%)\\
\midrule 
  \Checkmark& & &   51.7\\
 &\Checkmark & &  55.2\\
  & & \Checkmark&   62.1\\
\bottomrule
\end{tabular}
}
\end{table}
\textbf{The effect of CLIP-based initial balance:}
In the method section, the significance of the balance of initial data acquisition is discussed, and here we demonstrate it through experiments.
The results are shown in Table \ref{result_initial}, where random denotes that the initial samples are randomly selected.
The core set-based strategy is to extract the features of each region and then select the sample with the farthest distance according to the features.
The whole process can be divided into feature extraction, feature clustering, and selection.
To ensure fairness, for feature extraction, we abandon the backbone previously pre-trained on natural images such as ImageNet \cite{imagenet} and use an unsupervised training backbone specially used for aerial images, named seasonal contrast \cite{season}.
For feature clustering and selection, same with core set \cite{coreset}, we first divide all samples into multiple pools and then select the samples with the farthest distance in each pool.
The three methods in the table are consistent except for the different acquisition methods of initial data; others, like the subsequent data acquisition strategy and supervised contrastive learning, are the same.
As can be seen from the experimental results in Table \ref{result_initial}, if random initialization is used, even if a series of subsequent class balancing operations are used, the effect is just better than RALIS, SpAL, and CBAL, which fully demonstrates the significance of initialization data.
Compared with random, the core set-based method has an obvious improvement but still has an obvious gap with CLIP.
CLIP is also much faster than the core set-based method because clustering is not required.

\begin{table}[htbp]
\centering
  \caption{the effect of subsequent different balances}
  \label{result_following_balance}
    \scalebox{1.0}{
\begin{tabular}{c|c|c}
\toprule
data balance&performance balance &mIoU (\%)\\
\midrule 
 & &   34.4\\
  \Checkmark& &    49.9\\
 & \Checkmark&   62.1\\
\bottomrule
\end{tabular}
}
\end{table}
\textbf{The effect of performance-based balance for subsequent acquisition:}
In the subsequent labeled data acquisition, we use performance-based balance, verifying the results under different settings.
Except for the different strategies for subsequent labeled data, other experimental settings are the same as previous ones.
The results are shown in Table \ref{result_following_balance}.
It can be seen that if the subsequent balance is not considered, even if the network is trained with all other balances, the performance is not satisfied.
Class balance is important in active learning, and any absence will lead to the final imbalance, similar to the water in the bucket depending on the short board of the bucket.

\begin{table}[htbp]
\centering
  \caption{the effect of pseudo label}
  \label{result_pseudo}
    \scalebox{1.0}{
\begin{tabular}{c|c|c}
\toprule
pseudo label& balance of pseudo label&mIoU (\%)\\
\midrule 
 
 & &  56.3\\
 \Checkmark& &   55.5\\
 \Checkmark& \Checkmark&   62.1\\

\bottomrule
\end{tabular}
}
\end{table}
\textbf{The effect of pseudo label balance:}
The functions of pseudo labels are shown in Table \ref{result_pseudo}. If without the pseudo label, the effect will be reduced.
More importantly, if there is a pseudo label but no balance, the performance will be reduced even more.
This phenomenon is easy to understand. If we only use pseudo labels without balance, it is easy to classify confusing classes into the majority classes, thus resulting in errors.
By adding a different threshold to each class, the imbalance is suppressed, and the effect is significantly improved.

\begin{table}[htbp]
\centering
  \caption{the effect of supervised contrastive learning}
  \label{result_contrastive}
    \scalebox{1.0}{
\begin{tabular}{c|c|c}
\toprule
contrastive learning& balance &mIoU (\%)\\
\midrule 
 & &   52.6\\
 \Checkmark& &   57.4\\
 \Checkmark& \Checkmark&   62.1\\
\bottomrule
\end{tabular}
}
\end{table}
\textbf{The effect of feature extraction balance:}
For the first time, supervised contrastive learning is used to address class imbalances.
The results of the experiment are shown in Table \ref{result_contrastive}. It can be seen that contrastive learning improves performance significantly.
And the results are better with balance than without it. In practice, using balance only counts the loss of the classes with poor performance, while not using balance counts the contrastive loss of all classes.
In conclusion, the balanced contrastive loss is less computation but more effective.

\subsection{Influence of hyperparameters to the final result}
\label{sec_param}
\begin{figure}[h]
\centering
\includegraphics[scale=0.39]{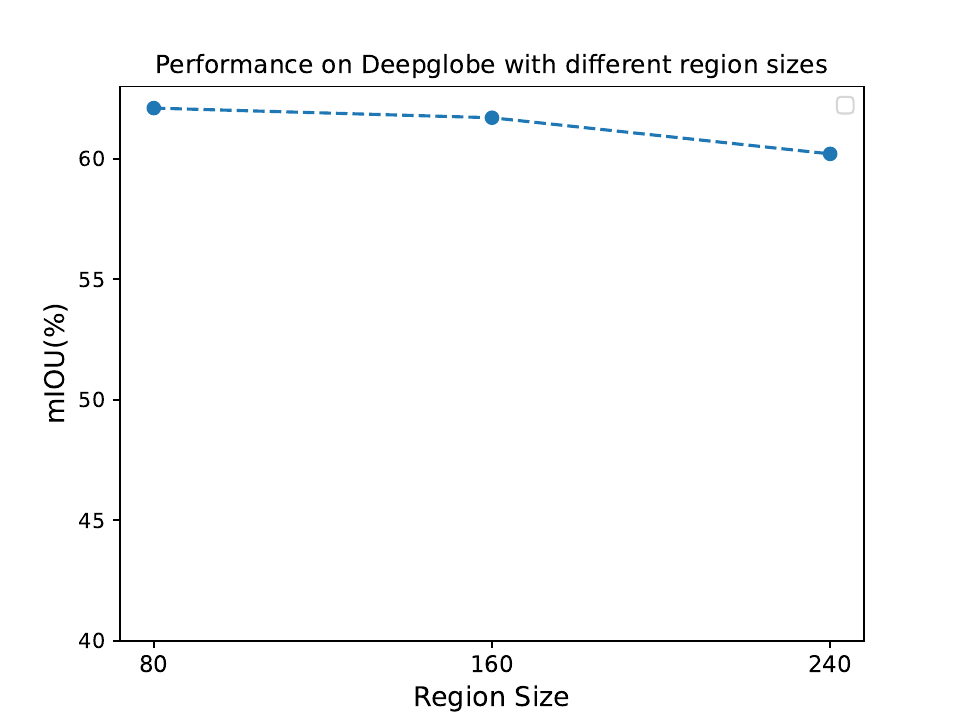}
\caption{
The performance change with different region sizes. 
}
\label{region_size}
\end{figure}
\textbf{The effect of region size:}
In the previous experiments, the default region size is $80 \times 80$.
In this chapter, we explore the impact of different region sizes on the final results.
We still use the most challenging Deepglobe dataset, use Deeplabv3 as the segmentation network, and select 20\% of all data for the label.
As shown in Fig. \ref{region_size}, region size impacts the final result.
Relatively, the smaller the region size, the better the result is.
This phenomenon can also be explained. \cite{spatial_div} demonstrates that labeled regions with a greater distance will produce better results.
With the same proportion of labeled pixels, the number of labeled regions will increase when using small regions, so the distance can be farther, and the result is improved.

\begin{table}[h]
\centering
  \caption{Results on Deepglobe with different coefficients}
  \label{result_number_anchors}
  \begin{tabular}{c|c}
    \toprule
Anchor Numbers& mIoU(\%)\\
\midrule
10&60.3\\
50&62.1\\
100&59.2\\
\bottomrule
\end{tabular}
\end{table}
\textbf{The effect of the number of anchor features:}
The influence of the number of anchor features ($T$ in Section \ref{sec_balance_contrastive}) on the result is shown in Table \ref{result_number_anchors}, and it can be seen that the number has little influence on the result, and when the value is set to 100, there is a slight drop.
This is also understandable. There are not adequate high-quality anchors, and when the number of anchors is 100, many features with low value will also be selected as anchors, resulting in a decline in results.
In general, it is better to choose about 50.

\begin{table}[h]
\centering
  \caption{Results with different edge detection methods}
  \label{table_edge_methods}
  \begin{tabular}{c|c}
    \toprule
Anchor Numbers& mIoU(\%)\\
\midrule
Canny&62.1\\
HED \cite{hed}&62.4\\
\bottomrule
\end{tabular}
\end{table}
\textbf{The effect of different edge detection methods:}
We use a network-based edge detection algorithm to compare with the traditional edge detection algorithm, and the results are shown in Table \ref{table_edge_methods}. It can be seen that the gap is not large. On the one hand, it shows that the edge detection of aerial images is not complicated, and on the other hand, it also shows the robustness of our method to edge detection methods.

\section{Conclusion}
Many aerial images are accessible with the improvement of sensors, but it is very difficult to label them at the pixel level. To solve this problem, we introduce an active learning method, and only a small amount of data need to be labeled to make the model obtain a very competitive performance.
We explore the most error-prone and uncertain regions in aerial images for segmentation tasks and propose a new edge-guided labeling unit that can select the most critical regions.
Meanwhile, we found that the imbalance between classes is the critical factor that seriously affected active learning performance, so we improved the imbalance in various aspects, including the balance of labeled data, the balance of pseudo labels, and the balance of feature extraction. 
The balance of labeled data is improved from two angles: initial labeled data and subsequent labeled data.
Through our improvement, active learning saves a lot of labeling costs in aerial images under the condition of ensuring performance.

\section{Acknowledgement}
This work is supported by NSFC Key Projects of International (Regional) Cooperation and Exchanges under Grant 61860206004, and NSFC projects under Grant 61976201.

\bibliographystyle{IEEEtran}
\bibliography{bibtex/bib/IEEEexample.bib}
\newpage
\begin{IEEEbiography}[{\includegraphics[width=0.8in,height=1in,clip,keepaspectratio]{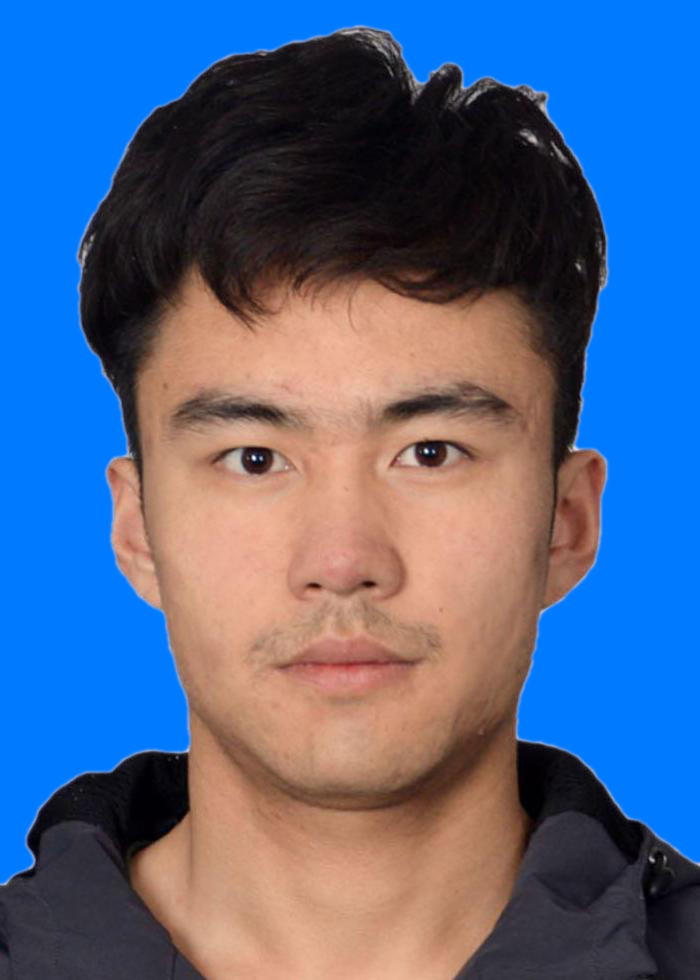}}]{Lianlei Shan} (Student Member, IEEE) is a Ph.D. student of School of Computer and Control Engineering in University of Chinese Academy of Sciences. He obtained his B.E. degree from Shandong University of Science and Technology in 2018. His research interests include remote sensing image processing and computer vision.
\end{IEEEbiography}
\begin{IEEEbiography}[{\includegraphics[width=0.8in,height=1in,clip,keepaspectratio]{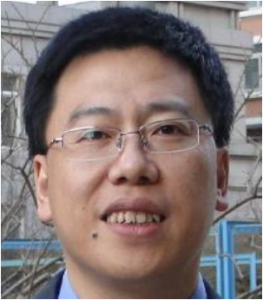}}]{Weiqiang Wang} (Member, IEEE) received the B.E. and M.E. degrees in computer science from Harbin Engineering University, Harbin, China, in 1995 and 1998, respectively, and the Ph.D. degree in computer science from the Institute of Computing Technology, Chinese Academy of Sciences (CAS), Beijing, China, in 2001. He is currently a Professor with the School of Computer and Controlling Engineering, University of CAS. His research interests include multimedia content analysis and computer vision.
\end{IEEEbiography}
\begin{IEEEbiography}[{\includegraphics[width=0.8in,height=1in,clip,keepaspectratio]{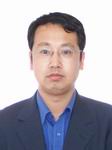}}]{Ke Lv} (Member, IEEE) received the B.S. degree from the Department of Mathematics, Ningxia University in 1993, and the master's and Ph.D degrees from the Department of Mathematics and Department of Computer Science, Northwest University in 1998 and 2003, respectively. He is currently a Professor with the University of Chinese Academy of Sciences. His research focuses on curve matching, 3-D image construction, and computer graphics.
\end{IEEEbiography}

\begin{IEEEbiography}[{\includegraphics[width=0.9in,height=1in,clip,keepaspectratio]{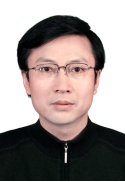}}]{Bin Luo} (Senior Member, IEEE) received his BEng. degree in electronics and MEng. degree in computer science from Anhui university of China in 1984 and 1991, respectively.
In 2002, he was awarded the Ph.D. degree in Computer Science from the University of York, the United Kingdom. He joined the University of York as a research associate from 2000 to 2004. 
\end{IEEEbiography}
\end{document}